# ARETE: an R package for Automated REtrieval from TExt with large language models

Vasco V. Branco 1,2,3, Jandó Benedek 4, Lidia Pivovarova 5, Luís Correia, 2, Pedro Cardoso 1,3

1. Laboratory for Integrative Biodiversity Research (LIBRe), Finnish Museum of Natural History Luomus, University of Helsinki, Pohjoinen Rautatiekatu 13, 00100 Helsinki, Finland

2. LASIGE and Departamento de Informática, Faculdade de Ciências, Universidade de Lisboa, 1749-016 Lisboa, Portugal

3. Centre for Ecology, Evolution and Environmental Changes (cE3c), Department of Animal Biology & CHANGE - Global Change and Sustainability Institute, Faculdade de Ciências, Universidade de Lisboa, 1749-016 Lisboa, Portugal

4. University of Veterinary Medicine, Department of Zoology, Budapest

5. University of Helsinki, Helsinki, Finland

## Abstract

1. A hard stop for the implementation of rigorous conservation initiatives is our lack of key species data, especially occurrence data. Furthermore, researchers have to contend with an accelerated speed at which new information must be collected and processed due to anthropogenic activity. Publications ranging from scientific papers to gray literature contain this crucial information but their data are often not machine-readable, requiring extensive human work to be retrieved.

2. We present the *ARETE R* package, an open-source software aiming to automate data extraction of species occurrences powered by large language models, namely using the chatGPT Application Programming Interface. This R package integrates all steps of the data extraction and validation process, from Optical Character Recognition to detection of outliers and output in tabular format. Furthermore, we validate ARETE through systematic comparison between what is modelled and the work of human annotators.

3. We demonstrate the usefulness of the approach by comparing range maps produced using GBIF data and with those automatically extracted for 100 species of spiders. Newly extracted data allowed to expand the known Extent of Occurrence by a mean three orders of magnitude, revealing new areas where the species were found in the past, which may have important implications for spatial conservation planning and extinction risk assessments.

4. *ARETE* allows faster access to hitherto untapped occurrence data, a potential game changer in projects requiring such data. Researchers will be able to better prioritize resources, manually verifying selected species while maintaining automated extraction for the majority. This workflow also allows predicting available bibliographic data during project planning.

# 1. Introduction

Spatial information, in particular those related with species occurrences, is essential for ecology, evolution, and conservation biology. Yet, such information is often scarce and biased, particularly for less-known taxa such as invertebrates, the so-called Wallacean shortfall (Cardoso *et al.* 2011). Although species occurrence data has been collected in widely accessible databases, including GBIF (GBIF, 2024) and, for citizen science data, iNaturalist (iNaturalist, 2024), much data remains hidden in unstructured text, including scientific and gray literature. As a result, it is used at a very slow pace or not at all, with each publication needing to be manually processed.

Generative AI (GenAI, see Supporting Information 1 for a glossary) is a powerful tool to extract, process and output information in novel ways. A wide selection of GenAI models have recently been made available to both professional and general audiences including proprietary initiatives such as chatGPT (OpenAI, 2024a), Claude (Anthropic, 2024) and Gemini (Gemini Team, 2023), and open source initiatives such as Llama (Touvron *et al.* 2023), Bloom (BigScience Workshop, 2023), and BERT (Devlin *et al.* 2019). GenAI uses are multiple, with Dasgupta *et al.* (2024) reporting a surge in research interest around it after 2017, in governance, architecture, drug discovery and many other fields.

GenAI also opened opportunities for (semi-)automated biodiversity data extraction. Some approaches have focused on determining the extent to which GenAI can be relied upon, including comparing the performance of different LLMs in data extraction and determining their factual knowledge and reasoning in ecological contexts (Castro *et al.* 2024; Dorm *et al.* 2025). More ambitious work set out to complement existing databases with automatically extracted data or synthesizing data based on abstracts (Rafiq *et al.* 2025; for specific cases see Scheepens *et al.* 2024; Bamba & Sato, 2025).

Past work was, however, specific to each question and not made available in ready-to-use software packages, limiting their adoption by the wider community. It also revealed some limitations, with their performance being often neglected or poorly explained. Specifically for geographical data, its evaluation tends to be mishandled, with most papers simplifying its representation by casting it as a binary, when these data are essentially non-binary variables.

Here we present a novel R package, Automated REtrieval from TExt (*ARETE),* that simplifies the automatic extraction of species occurrence data from unstructured text. Our goals are to: 1) detail the typical workflow of *ARETE*; 2) describe the process of fine-tuning models; 3) assess model performance using both default and fine-tuned models; 4) test its capability on literature using a sample of 100 species of spiders randomly selected across the world. We release this package with the goal of allowing researchers to speed up the process of extracting occurrence data for all taxa, this way streamlining biodiversity-related projects that rely on it.

# 2. The *ARETE* package

## 2.1. User input

*ARETE* is an *R* and *Python* based pipeline (Figure 2.1) for the deployment of large language models (LLM) to extract occurrence data from unstructured text, with data from the *openai Python* API being processed and delivered through an *R* package, the most used language by ecologists and conservation biologists. Accessing LLMs directly through an Application Programming Interface (API) can be taxing for non-experts. In contrast, the basic *R* function in *ARETE*, *get_geodata()* only requires the following parameter inputs:

1) *path*, string pointing to a file with species data in either *.pdf* or *.txt* format, e.g: *./folder/file.pdf*;
2) *tax*, string designating the species to search for in the given document (if set to NULL all data will be retrieved);
3) *user_key*, list with two elements: *key*, a string with the user's API key and *premium*, a Boolean setting the type of user key when applicable for the chosen *service*. Both free keys and premium keys are allowed, with keys from free / trial services having drawbacks such as smaller context windows, which reduce the model's performance;
4) *service*, optional string designating the service to be used, default "*GPT*". Right now, only OpenAI's service, chatGPT, is available;
5) *model,* optional string designating the model to be used. As of publication date we recommend the options "*gpt-3.5*" (default) or "*gpt-4o*", as these select a tested model instance within that version (for the full model documentation please consult OpenAI, 2024c);
6) *outpath*, optional string pointing to a path to save output to, default is NULL, not saving;
7) *verbose*, optional Boolean determining if output should be printed, default is TRUE.

**FIG 2.1**. Framework and workflow for *ARETE*. Top row boxes (a) give information on each step of the workflow, while the bottom row (b) contains the data retrieved at the end of each step. In this example, a user supplies text containing fictitious data for two species, *Araneus diadematus* and *Arctosa alpigena*, collected in Finland and Sweden. Unbeknownst to the user, one of the localities for *A. diadematus* is incorrect and contains coordinates for Kilpisjärvi (b.1). The text is processed and any invalid characters, such as sex signs, are replaced (b.2). The information is extracted and presented as an analysis-ready table (b.3). The user runs the optional outlier detection step and is warned that one of the data points is a likely outlier due to its distance in spatial and environmental space (b.4).

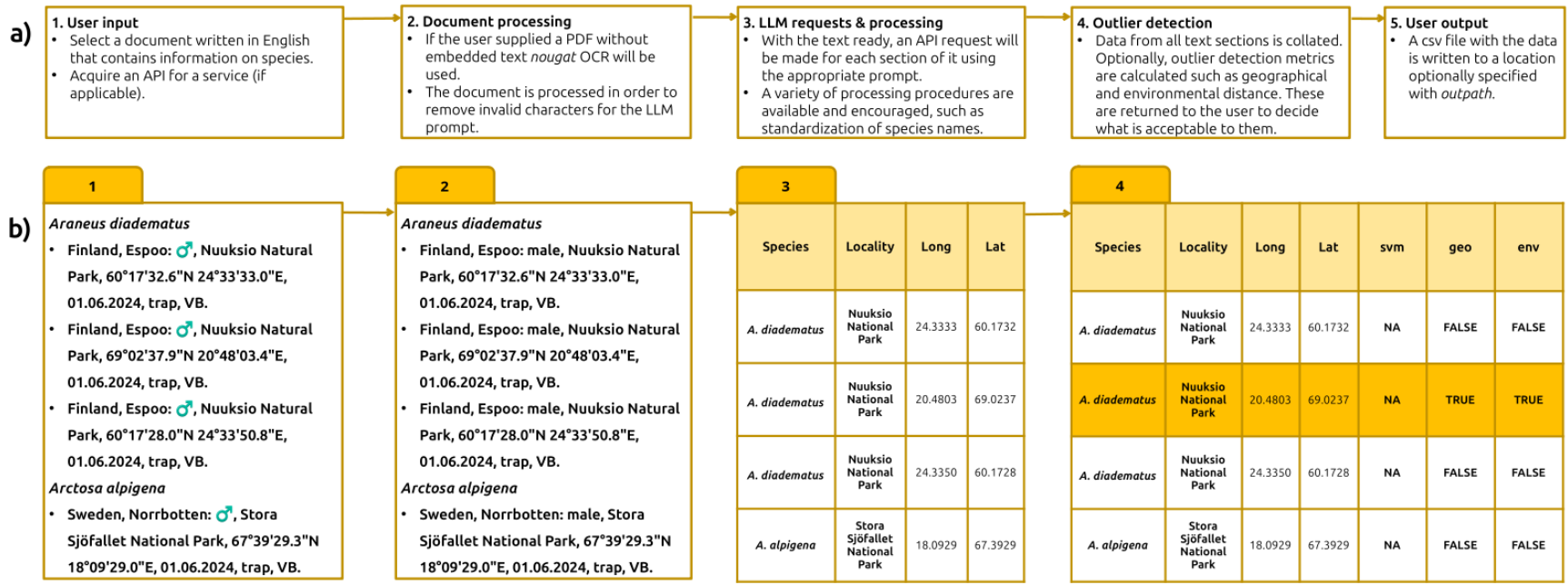

## 2.2 Document processing

A text format file can be directly used by the package. If the input is in *.pdf* format its embedded text is extracted. If no text is embedded, the user is notified that *nougat* OCR (Blecher *et al.* 2023), an academic document parser based on *Tesseract* OCR (Smith, 2007), will be used. The extracted text is processed, removing invalid characters.

## 2.3 LLM requests

The text is broken in one or more parts depending on the characteristics of the text (*path*), *model* and *service* given (See section 2.1.1, "User input"). This is done to comply with existing maximum token limits, which will vary with *model* and *service*. Each part of the original text is then appended to a prompt that is best suited for that service (see Supporting Information 2 for details).

Most proprietary LLM offer both free services and premium services that offer a higher rate of requests and/or characters per prompt. Both are taken into consideration and can be selected by the user. For development and testing of the package we used models "*gpt-3.5-turbo-1106*" and "*gpt-4o-mini-2024-07-18*". Although there are differences between different instances of the same model version these should not prevent their use. Support of additional models and services, including open-source models is planned (See section 6. "Availability & future updates").

## 2.4 Outlier processing

As an optional step after data extraction, *ARETE* uses the function *detect_outlier()* of package *gecko* (Branco *et al.*, 2023). It flags possible outliers using geographical distances (*geo*), environmental distances (*env*), and support vector machines (*svm*). Environmental distance is calculated as the euclidean distance in an *n*-dimensional space where every dimension is a normalised environmental feature, namely climate layers from WorldClim (WorldClim, 2024; Fick & Hijmans, 2017). Support vector machines (SVM) are implemented as in Liu *et al.* (2017), representing pseudo-Species Distribution Models (SDM) derived from two-class presence-only SVMs also using WorldClim data. This model is capable of using pseudo-absence points, implemented with the *ksvm()* function in the R package *kernlab* (Karatzoglou *et al.*, 2024). In place of pseudo-absence points, users are encouraged to supply true absences when available to increase the accuracy of results. For both *geo* and *env*, all distance values above a given threshold (95% as default) are flagged as potential outliers (*i.e.*, points significantly different from every other point). The SVM method identifies outliers as those points falling outside the environmental envelope given by the multidimensional vectors. The full collection of metrics is offered by default and action is not taken automatically - the user determines how conservative they want to be.

## 2.5 Model validation

*ARETE* also provides model validation options using function *performance_report()*. This takes as arguments two matrices with species and localities data, one being the output of *get_geodata()* and the other a user-supplied matrix for ground-truth. Accuracy, recall, precision and F1 are calculated for coordinates along with each of these metrics inversely weighted by the mean geographical distance of that data point to all others. This way, large errors in coordinates have a proportional effect on the metric (similar to the outlier checking process described in section 2.1.4).

For locality data *ARETE* calculates the mean minimum Levenshtein distance between locality strings (Figure 2.3), a metric often used in fuzzy matching problems (Kalyanathaya, Akila & Suseendren 2019). This metric is the minimum number of character editions required to go from one string to another, using character changes, additions and deletions. All metrics are presented separately rather than combined into one overall performance metric. The results of the report are saved to *.Rmd* files, one for each species and one global report.

**FIG 2.3**. During validation, the text extracted by the LLM is compared to user-supplied reference data. In a text with two terms, a) and b), the Levenshtein distance is calculated between each term and every other LLM term (Step 1), keeping the minimum value of these distances (Step 2). The mean of these values becomes the performance score for the entire document (Step 3).

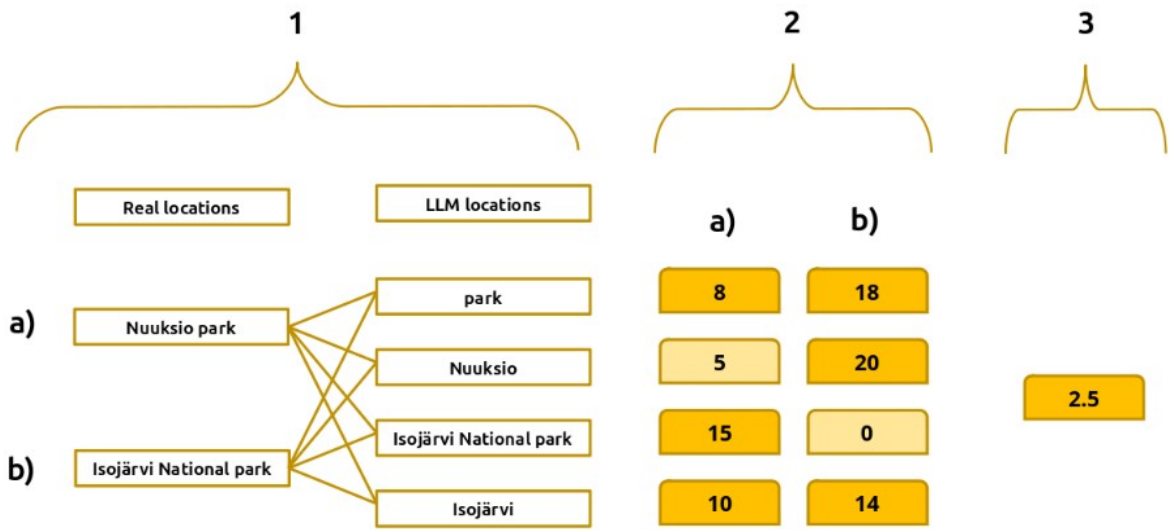

## 3. Models

In order to test *ARETE*'s capabilities (See section 2, "The ARETE package") we used *RECODE*, a corpus of manually annotated ecological and taxonomic papers (Branco *et al. in prep*). In *RECODE*, annotated papers are marked with labels of interest and their corresponding relational data, such as the name of a species and the localities relating to it, making them ideal for both training and validation of all *ARETE* models. Furthermore, most entries in *RECODE* focus on insects (Class Insecta) and spiders (Order Araneae), two mega-diverse taxa that would benefit the most from automated data extraction. Spiders alone make up close to 52,000 species (WSC, 2024), nearly double the amount of mammal species (21,639), with both groups being dwarfed by the amount estimated for insects (1,114,071; GBIF, 2024).

The models used were "*gpt-3.5-turbo-1106*" and "*gpt-4o-mini-2024-07-18*", both in their standard versions and after fine-tuning. LLMs being complex models are usually trained in generic data which can leave their performance below expectations for very specific tasks. In order to mitigate this issue, they can be fine-tuned, *i.e.* perform a secondary training process on new data tailored to a specific problem the user wants solved. In order to determine how fine-tuning LLMs can improve performance and how good of an investment this process is for prospective users, we trained several fine-tuned versions of chatGPT. Fine-tuning of GPT models was achieved through usage of both the GPT browser UI and the *openai* Python library (OpenAI, 2024b). Both are paid models available in the OpenAI API.

## 4. Model performance

### 4.1 With no fine-tuning

We evaluated the performance of the standard GPT model "*gpt-3.5-turbo-1106*" with no fine-tuning by comparing the data extracted by ARETE for 50 papers present in RECODE (Table 4.1.1, Table 4.1.2). Due to the special importance of False Positives (including hallucinations) these were manually checked.

**TABLE 4.1.1**. Confusion matrix for a comparison between records from automatically extracted data and an annotated corpus.

|          |       | Predicted |       |
|----------|-------|-----------|-------|
|          |       | TRUE      | FALSE |
| Observed | TRUE  | 110       | 34    |
|          | FALSE | 10        | *NA*  |

**TABLE 4.1.2**. Performance metrics. These values are part of a larger report available, along with its code, on the project's repository and R package.

| Metric | Value |
| --- | --- |
| Accuracy | 0.714 |
| Recall | 0.764 |
| Precision | 0.917 |
| F1 | 0.833 |

Out of 10 False Positives, we found that half were not the model's fault, and were due to misinterpretation of characters by OCR. This happens in either the species name or coordinate fields (also a common contributor to False Negatives). Excluding such errors leaves only 5 legitimate model "hallucinations", essentially due to two types of error. The first is misinterpreting non-interest species, such as those in references, as of interest. This is the most common type of error and that is the reason for the prompt used in the LLM to explicitly remove it (See Supporting Information 2). The second type of error is misattributing the localities of one species to another valid species in the text. This error is very infrequent and should often be detected by the outlier processing abilities of *ARETE*.

## 4.2 With fine-tuning

We followed with a *k-fold* learning process. Due to the challenge of classifying and attributing performance to a variable string outcome we performed a manual evaluation of the fine-tuning process, where the contribution of each entry to the confusion matrix (and following performance metrics) was weighted by human interpretability of locality and coordinate data (Table 4.1.1, Table 4.2.2, see Supporting Information 3).

**TABLE 4.2.1**. Confusion matrix for the best performing model within standard deviation across all folds and series from automatically extracted data and an annotated corpus.

|  |  | Predicted | |
|---|---|---|---|
|  |  | TRUE | FALSE |
| **Observed** | TRUE | 35 | 2 |
|  | FALSE | 1 | NA |

**TABLE 4.2.2**. Performance metrics. These values are part of a larger report available, along with its code, on the project's repository and R package.

| Metric | Value |
|---|---|
| Accuracy | 0.921 |
| Recall | 0.972 |
| Precision | 0.945 |
| F1 | 0.958 |

## 5. Use case: extracting spider data

On the 22nd of January 2024 we sampled 100 random species from the 52000+ recorded in the World Spider Catalog's (WSC) Daily Species Export file (*species_export_20240122.csv*) and downloaded every *.pdf* file associated with each species. This resulted in a total of 321 files downloaded. Of these, 155 were removed as their main language was not English. For extraction, we used our best performing GPT4o fine-tuned model (fold I and 20 data points) as it had the lowest Levenshtein distance across all models.

Additionally, we analysed the spatial distribution of data obtained from GBIF as it has been found before to be spatially biased. Dasgupta *et al.* (2024) report that in an analysis of arthropods of small ranges (<= 25 km x 25 km), 30 countries out of 78 were identified to hold 78.2% of species. Likewise, from a sample of 100 species, 60 of which with data in GBIF, we found that out of 58 countries, 30 countries represented 74.28% of species occurrences.

In order to ensure the extraction of spider data was consistent with the model performance described previously, 50 of the 100 species were selected for manual evaluation. As taxonomic papers vary greatly in quality (*i.e.*: OCR derived errors, data structure, and presence of non-english text) due to the age and size of some of the sources, a two-phase search was conducted, the first using a standard GPT search, the second manually. We found that performance was lower but overall consistent with the results previously found, with most of the errors being False Negatives (Table 5.2). Additionally, most of the errors found were centered around a few problematic papers that due to a variety of issues (e.g: errors in embedded text in *.pdf*, possibly due to prior OCR errors) heavily skewed performance results. The three most error prone papers were responsible for 51.5% of False Negatives despite being less than 5% of our sample.

**TABLE 5.2**. Performance metrics for records extracted by ARETE in a selection of taxonomic papers from a sample of 50 spider species.

| Metric | Value |
| --- | --- |
| Accuracy | 0.544 |
| Recall | 0.607 |
| Precision | 0.840 |
| F1 | 0.704 |

The sources associated with 100 species from our sample were analysed to assess how the data extracted complemented existing GBIF data (Fig 5.1). An exploratory analysis revealed that data extracted by *ARETE* referenced most species in the sample (67%), higher than what is present in GBIF (60%) with significant overlap (44%) between the two (See Supporting Information 4 for details). This leaves only a total of 17% that could be found in neither.

**FIG 5.1.** Data retrieved from GBIF (yellow, right) and automatically extracted with *ARETE* (blue, left). Species were chosen from a random sample of 100 spider species listed in the World Spider Catalog (WSC).

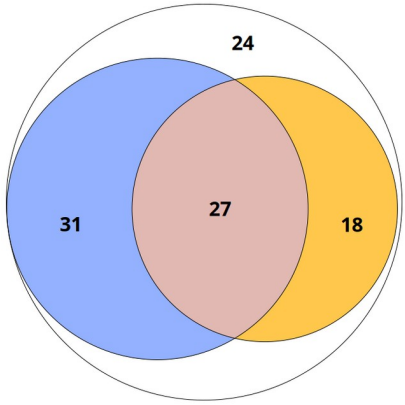
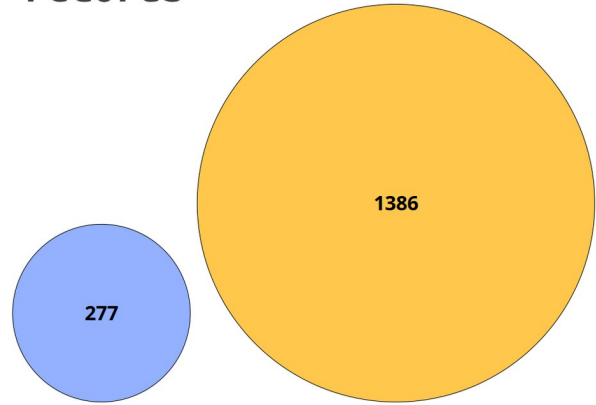
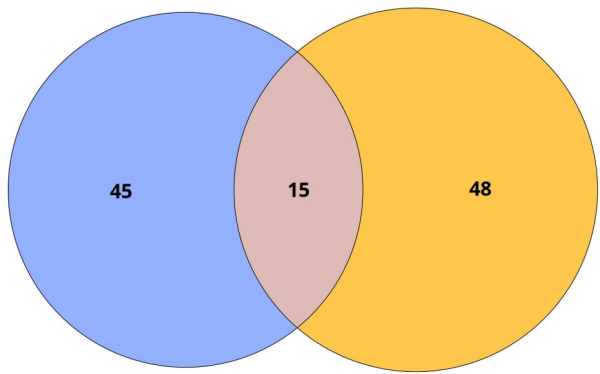
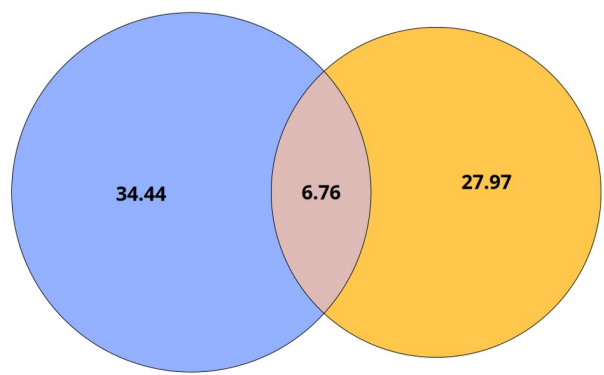

There is a drop in the amount of species found by *ARETE* when considering only those entries that contain longlat data. However, this is to be expected. For the 50 species selected for manual validation only 62% of entries had longlat information, mirroring this situation.

Analysing how *ARETE* might change current species distributions, we found that the vast majority of countries common to the GBIF and *ARETE* datasets were in Eurasia (Fig. 5.2).

**FIG 5.2**. Number of species per country within our random sample of 100 spider species for data obtained from GBIF (a), ARETE (b), and those recorded for the first time with ARETE (c).

a)
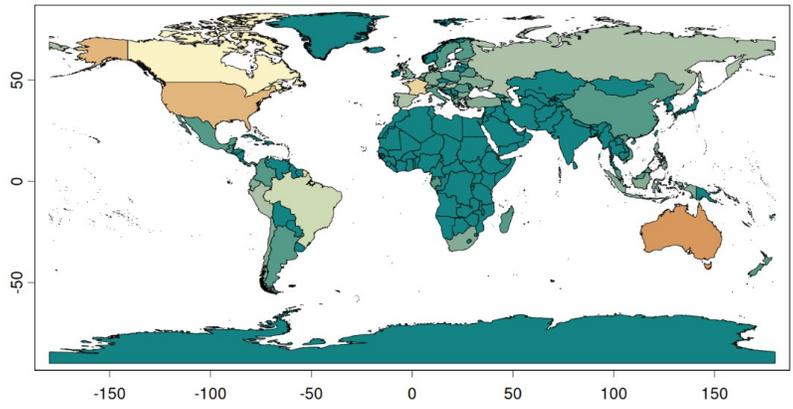

b)
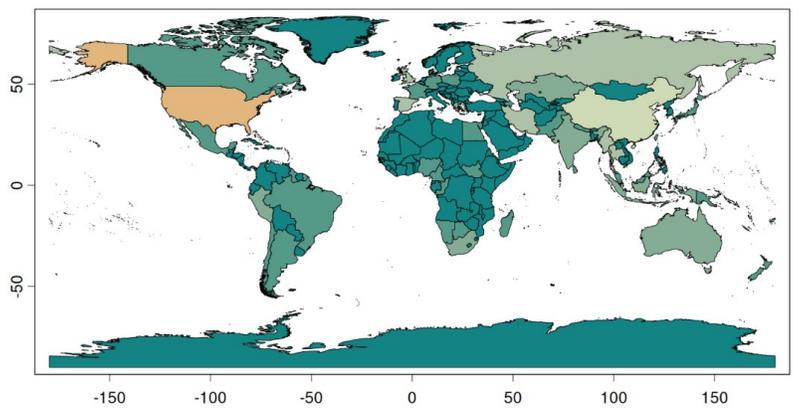

c)
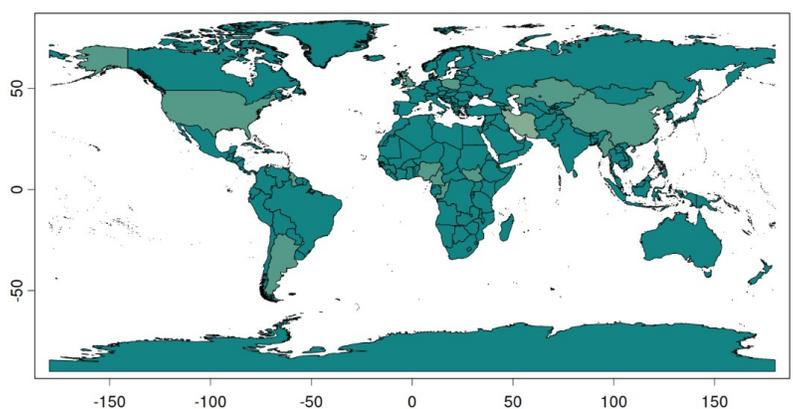

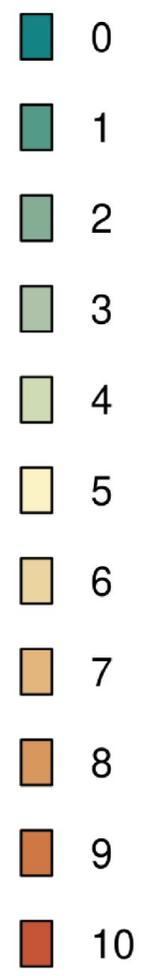

Lastly, we looked at the potential contributions of *ARETE* for each species' Extent of Occurrence (EOO), calculated as the area encompassed by the minimum convex polygon which contains all occurrence points. For the 22 species with geospatial data in common between GBIF and *ARETE* (Fig 5.1), there was a mean three orders of magnitude increase in EOO (*i.e*: exactly, 1949.84 times increase). Additionally, in 9 of these species, these increases in EOO were sufficient to change the IUCN threat status using criterion B, highlighting how impactful the newly extracted data can be. This also aligns with previously known limitations in data, as Shirey *et al.* (2019) reported hypothetical IUCN red list classifications for endemic iberian spider species to be composed of 93.1% Data Deficient (DD), substantially worse than the 41.9% DD obtained from an extensive bibliographic review.

## 6. Availability & future updates

*ARETE* is meant to be continuously updated and we have a collection of features planned. These include adding support for additional LLMs, in particular open-source alternatives (Llama, Bard, Palm, Gemini, DeepSeek, etc.) and expanding the package's scope to also encompass the extraction of trait data. We also plan on, as much as possible, adding taxonomic referencing to disentangle synonyms in text and address current limitations, which are discussed in full in Supporting Information 5. Additionally, the full list of papers used for fine-tuning and their corresponding metadata can be consulted in Supporting Information 6.

As with most R packages, our master branch is available from:

- https://cran.r-project.org/package=arete
- https://github.com/VascoBranco/arete

And we encourage users to submit their feedback at:

- https://github.com/VascoBranco/arete/issues

## Acknowledgments

We thank the University of Helsinki's Global Change and Conservation research group for their feedback, especially to Karolina Lukasik for her suggestions on measuring model performance. We would also like to thank IIris Lahin and Lenka Baranovičová for their work in annotating our fine-tuning corpus. V.V.B. was supported by Kone Foundation, Finland. L.C. was supported by FCT, Portugal through LASIGE Research Unit, ref. UID/00408/2025-LASIGE. P. C. has received funding from the European Union's research and innovation programme Horizon Europe (BioMonitor4CAP, grant agreement #101081964), cE3c (https://doi.org/10.54499/UIDB/00329/2020), and CHANGE (https://doi.org/10.54499/la/p/0121/2020).

## References

- Anthropic (2024) 'The Claude 3 Model Family: Opus, Sonnet, Haiku'. https://www-cdn.anthropic.com/de8ba9b01c9ab7cbabf5c33b80b7bbc618857627/Model_Card_Claude_3.pdf (Accessed: October 2024).

- Bamba, M. & Sato, S. (2025) 'Expanding plant trait databases using large language model: A case study on flower color extraction.' *bioRxiv*, 2025.02.11.637746.

- BigScience Workshop (2023) 'BLOOM: A 176B-Parameter Open-Access Multilingual Language Model', *arXiv*. doi: 10.48550/arXiv.2211.05100.

- Blecher, L., Cucurull, G., Scialom, T. & Stojnic, R. (2023) 'Nougat: Neural Optical Understanding for Academic Documents', *arXiv*. doi: 10.48550/arXiv.2308.13418.

- Branco, V.V., Cardoso, P. & Correia, L. (2023) 'gecko: Geographical Ecology and Conservation Knowledge Online'. v.1.0.1. CRAN, https://cran.r-project.org/web/packages/gecko/index.html

- Branco, V.V., Pivovarova, L., Lintulaakso, K., Correia, L.M., Baranovičová, L., Lahin, I., Dias, F., Filipe, D. & Cardoso, P. (in prep). 'RECODE - Relational Ecological COrpus for Data Extraction'.

- Cardoso, P., Erwin, T.L. Borges, P.A.V. & New, T.R. (2011) 'The seven impediments in invertebrate conservation and how to overcome them', *Biological Conservation*, 144(11). doi: 10.1016/j.biocon.2011.07.024.

- Castro, A., Pinto, J., Reino, L., Pipek, P. & Capinha, C. (2024) 'Large language models overcome the challenges of unstructured text data in ecology'. *Ecological Informatics*, 82, 102742. doi: 10.1016/j.ecoinf.2024.102742.

- Dasgupta, S., Blankespoor, B. & Wheeler, D. (2024) 'Toward Better Conservation: A Spatial Analysis of Species Occurrence Data from the Global Biodiversity Information Facility', *Earth Syst. Sci. Data Discuss*. [preprint, in review]. doi: 10.5194/essd-2024-241

- Devlin, J., Chang, M., Lee, K. & Toutanova, K. (2019) 'BERT: Pre-training of Deep Bidirectional Transformers for Language Understanding'. In: *Proceedings of the 2019 Conference of the North American Chapter of the Association for Computational Linguistics: Human Language Technologies*, Volume 1 (Long and Short Papers), (pp. 4171–4186). doi: 10.18653/v1/N19-1423

- Dorm, F., Millard, J., Purves, D., Harfoot, M. & Aodha, O.M. (2025) 'Large language models possess some ecological knowledge, but how much?', *bioRxiv*. doi: 10.1101/2025.02.10.637097

- Fick, S.E. & Hijmans, R.J. (2017) 'WorldClim 2: New 1-km spatial resolution climate surfaces for Global Land Areas', *International Journal of Climatology*, 37(12), 4302–4315. doi: 10.1002/joc.5086.

- GBIF (2024) https://www.gbif.org/ (Accessed: 16 December 2024).

- Gemini Team (2023) 'Gemini: A Family of Highly Capable Multimodal Models', *arXiv*. doi: 10.48550/arXiv.2312.11805.

- Gupta, P., Ding, B., Guan, C. & Ding, D. (2024) 'Generative AI: A systematic review using topic modelling techniques', *Data and Information Management*. doi: 10.1016/j.dim.2024.100066.

- *INaturalist* (2024). https://www.inaturalist.org/ (Accessed: 16 December 2024).

- Kalyanathaya, K.P., Akila, D. & Suseendren, G. (2019) 'A Fuzzy Approach to Approximate String Matching for Text Retrieval in NLP', *Journal of Computational Information Systems*, 15(3), 26–32.

- Karatzoglou, A., Smola, A., Hornik, K., Maniscalco, M.A., Teo, C.H. (2024) 'kernlab: Kernel-Based Machine Learning Lab'. CRAN, https://cran.r-project.org/web/packages/kernlab/index.html.

- Liu, C., White, M. & Newell, G. (2017) 'Detecting outliers in species distribution data', *Journal of Biogeography*, 45(1), 164–176. doi: 10.1111/jbi.13122.

- OpenAI (2024a) 'GPT-4 Technical Report', *arXiv*. doi: 10.48550/arXiv.2303.08774.

- OpenAI (2024b) 'Openai/openai-python: The official Python Library for the openai API', GitHub. v.1.13.3. https://github.com/openai/openai-python (Accessed: December 2024).

- OpenAI (2024c). 'OpenAI API - Models'. https://platform.openai.com/docs/models (Accessed: October 2024).

- Rafiq, K., Beery, S., Palmer, M.S., Harchaoui, Z. & Abrahms, B. (2025) 'Generative AI as a tool to accelerate the field of ecology', *Nature Ecology & Evolution*, 9, 378-385. doi: 10.1038/s41559-024-02623-1.

- Scheepens, D., Millard, J., Farrell, M. & Newbold, T. (2024) 'Large language models help facilitate the automated synthesis of information on potential pest controllers', *Methods in Ecology and Evolution*, 15, 1261-127. doi: 10.1111/2041-210X.14341.

- Shirey, V., Seppälä, S., Branco, V.V. & Cardoso, P. (2019). 'Current GBIF occurrence data demonstrates both promise and limitations for potential red listing of spiders', *Biodiversity Data Journal*, 7, 47369. doi: 10.3897/BDJ.7.e47369.

- Smith, R. (2007) 'An Overview of the Tesseract OCR Engine'. https://tesseract-ocr.github.io/docs/tesseracticdar2007.pdf (Accessed: October 2024).

- Touvron, H., Lavril, T., Izacard, G., Martinet, X., Lachaux, M., Lacroix, T., Rozière, B., Goyal, N., Hambro, E., Azhar, F., Rodriguez, A., Joulin, A., Grave, E. & Lample, G. (2023) 'LLaMA: Open and Efficient Foundation Language Models', *arXiv*. doi: 10.48550/arXiv.2302.13971.

- World Spider Catalog (2025) 'World Spider Catalog'. Version 25.5. Natural History Museum Bern. doi: 10.24436/2 (Accessed: January 2025).

- WorldClim (2024). https://www.worldclim.org/ (Accessed: January 2025).

# SUPPORTING INFORMATION - 1

## S.1.1 Glossary

**AI.** *Artificial intelligence*. Field of data science that studies "*[...] agents that receive precepts from the environment and perform actions. Each such agent implements a function that maps percepts to actions*", with these functions exemplified as "*reactive agents, real-time planners, decision-theoretic systems, and deep learning systems*" (Russel & Norvig, 2021).

**ML.** *Machine learning*. Sub-field of AI that seeks to develop software capable of solving learning problems. Data scientist Tom Mitchell defined 'a well-posed learning problem', or a 'machine learning' problem, as when 'a computer program is said to learn from experience E with respect to some task T and some performance measure P, if its performance on T, as measured by P, improves with experience E' (Mitchell, 1997).

**LLM.** *Large Language Model*. A machine learning model that uses neural networks and a self-attention process, assigning different attention priority values, *i.e.*, an order of importance, to each token. These tokens are turned into vectors in a high dimensionality space, the value of which being determined by a vast training process that analyses their conceptual relationships in text (Vaswani *et al.* 2017).

**API**. *Application programming interface*. A set of definitions in a computer language that a software system makes available so that its functionalities can be used by other software systems.

**EOO**. *Extent of Occurrence*. Defined as *"the area contained within the shortest continuous imaginary boundary which can be drawn to encompass all the known, inferred or projected sites of present occurrence of a taxon, excluding cases of vagrancy"* (IUCN, 2024).

**Equation 1**. Precision:

$$Precision = \frac{TP}{TP+FP},$$

with TP amount of True Positives and FP amount of False Positives

**Equation 2**. Recall:

$$Recall = \frac{TP}{TP+FN},$$

with TP amount of True Positives and FN amount of False Negatives

**Equation 3**. F1. The harmonic mean of precision and recall, attempting to balance these two measures.

$$F1 = \frac{2 * Precision * Recall}{Precision + Recall}$$

**Disclaimer**: The three metrics above are sensitive to class imbalance. This means that their results will depend on which class is considered "positive".

## S.1.2 References

- IUCN (2024). Guidelines for the application of IUCN Red List of Ecosystems Categories and Criteria, Version 2.0. Keith, D.A., Ferrer-Paris, J.R., Ghoraba, S.M.M., Henriksen, S., Monyeki, M., Murray, N.J., Nicholson, E., Rowland, J., Skowno, A., Slingsby, J.A., Storeng, A.B., Valderrábano, M. & Zager, I. (Eds.). Gland, Switzerland: IUCN. doi: 10.2305/CJDF9122

- Mitchell, T.M., 1997. Machine learning, 1st ed. McGraw-Hill Education. ISBN: 978-0071154673.

- Russel, S.J. & Norvig, P. (2020). Artificial Intelligence: A Modern Approach (4th ed.). Pearson Education. ISBN: 978-0-13-461099-3.

- Vaswani, A., Shazeer, N., Parmar, N., Uszkoreit, J., Jones, L., Gomez, A.N., Kaiser, L. & Polosukhin, I. (2017). Attention is All you Need, arXiv. doi: 10.48550/arXiv.1706.03762

# SUPPORTING INFORMATION - 2

The following is the prompt used for GPT API requests in *ARETE*.

"*Read the following document then reply only a table consisting of three columns, Species, Location, and Coordinates containing new geographic data and the in-text coordinates of species mentioned in the following document. Include locations as they are written in the text, aside from the special character: |. Include locations without coordinates. Do not include species with neither location nor coordinate data. Do not include those in scientific articles or outside references:*"

As the first step, the model is asked to reply with nothing but a table consisting of three columns, species, location, and coordinates containing new geographic data and the in-text coordinates of species mentioned in the document.

We specify that locations are to be included in the table as they show written in the text. GPT has issues fulfilling this request when instructed but makes no attempt to do so unless specifically reinforced in this way despite potentially being assumed at the start. We make an exception for the special character "|" which greatly improves output reliability as this character is used to create the output table.

We give specific instructions over the kind of data to be included. We request the inclusion of locations without coordinates as GPT will often not include these unless specified in the prompt. We also request that species with neither kind of info be excluded as these are frequent (passing references to a species in a sentence, for example) and consume resources. Finally, in order to also cut back on the amount of passing references incorrectly picked up by the model we also request that scientific articles or outside references are to be excluded. It must be noted however, that due to the nature of LLMs this is not always respected in the model's output.

# SUPPORTING INFORMATION - 3

The fine-tuning followed a *k*-fold structure training process where from an initial set of 100 papers, 5 folds (A-E) of 20 papers each were created. One fold was then selected as the test data and the remaining folds were used as a pool for training, being sampled depending on the amount of training desired (10, 20, 40, 80 and 0 for no fine-tuning).

Due to the challenge of classifying and attributing performance to a variable string outcome we decided to perform a manual evaluation of the fine tuning process. In this evaluation the standard F1 score formula (see Equation 3 in the Glossary) is adapted to use True Positive, False Positive and False Negative classifications that are the product of a mean between a binary score (for coordinates, are they there or not) and a gradient score determining text closeness that is determined through human interpretation, which followed a few guidelines.

**TABLE S.3.1**. Guidelines used in scoring the performance of locations extracted by GPT. Should a term present in annotated data not be present in any way in the data suggested by GPT, then the data is instead labelled as a False Negative with a weight always of 1.

| E.g | Cidade Universitária, Alameda da Universidade, Lisboa, Portugal |
|---|---|

| Score | Guideline | Example |
|---|---|---|
| 1 | Total or near total match between terms, accounting for minor differences such as punctuation or prepositional words that do not reduce meaning. | cidade universitaria alameda da universidade lisboa portugal |
| 0.75 | Incomplete locations that do not significantly reduce specificity, often by only missing the smallest territorial entities, e.g. a parish or street. | Alameda da Universidade, Lisboa, Portugal |
| 0.5 | Incomplete locations that do not significantly reduce specificity, but are quite different from the original text. | Alameda da Universidade |
| 0.25 | Very incomplete locations that significantly reduce specificity. Usually, these describe a large territorial entity and are scored instead as False Negatives, as the information to be extracted would not be useful. | Portugal |

These scores are attributed in two spreadsheets, one containing the data known to be in a given paper (*obs*), the other (*pred*) containing the data extracted by GPT (Table S.3.1). Scores are then combined according to what is described under Equation S.3.1. In order for the same True Positives to not be counted several times, their weights are set to 0 in *pred*.

**TABLE S.3.2**. Mock example of weighted F1 score. Locations (*loc*) and coordinates (*coord*) are scored (*loc_score*, *coord_score*) according to their proximity to each other. Scores are combined towards a *classification* that has a given *weight*.

| Data in paper | | | | | |
|---|---|---|---|---|---|
| **loc** | **coord** | **loc_score** | **coord_score** | **classification** | **weight** |
| Lisbon | 38.85, -8.96 | 0 | 0 | FN | 1 |
| Algarve, Fuseta | 37.03, -7.75 | 0.5 | 1 | TP | 0.75 |

| Data suggested by LLM | | | | | |
|---|---|---|---|---|---|
| **loc** | **coord** | **loc_score** | **coord_score** | **classification** | **weight** |
| Helsinki | 60.15, 24.95 | 0 | 0 | FP | 1 |
| Algarve | 37.03, -7.75 | 1 | 1 | TP | 0 |

**EQUATION S.3.1**. Formula for combining proximity scores.

$$Score(obs) = \begin{cases} FN, weight = 1 & \text{if } x = 0 \wedge y = 0. \\ TP, weight = \frac{x+y}{2} & \text{otherwise.} \end{cases} \quad (1)$$

$$Score(pred) = \begin{cases} FP, weight = \frac{2-(x+y)}{2} & \text{if } x > 0 \vee y > 0. \\ TP, weight = 0 & \text{otherwise.} \end{cases} \quad (2)$$

With *x* and *y* being the location and coordinate proximity scores.

Due to the large amount of data, checking each row was simplified by preprocessing all rows in R. We calculated the levenshtein distance between each coordinate and location in our test data and every location and coordinate suggested by GPT. This way we had at hand the closest strings suggested by GPT and whether or not these closest terms came from the same row, which was highly predictive of incongruences.

**FIG S.3.1.** Performance of "*gpt-3.5-turbo-1106*" at extracting species location data (coordinates and name) calculated as the weighted F1 (b) and the mean minimum Levenshtein distance (a) between each point suggested by GPT and every other point from our ground truth dataset. The fine-tuned models used are trained on different folds (A-E) and differing amounts of data (x-axis). The black line is the mean and standard deviation of all folds.

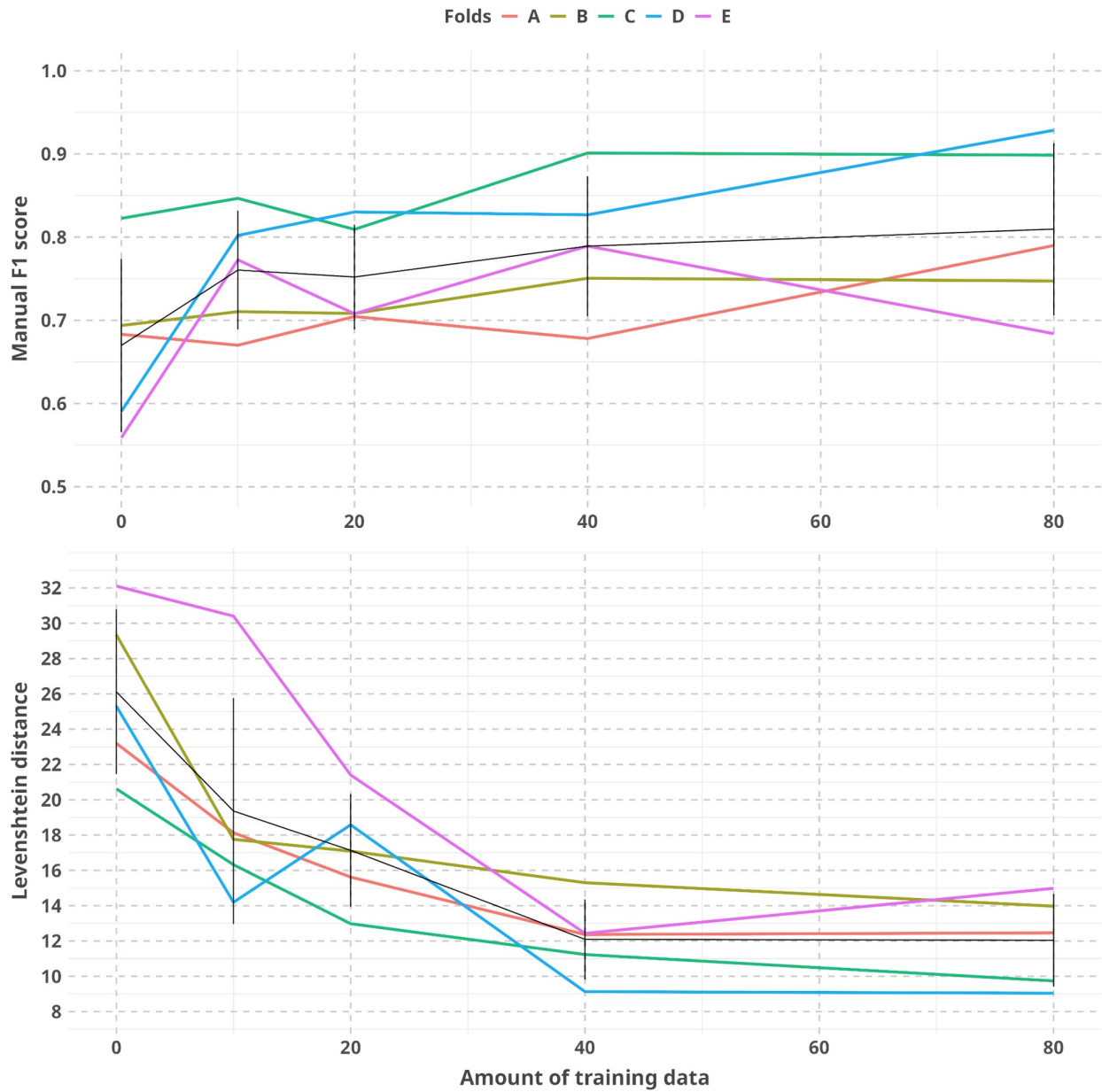

After manually evaluating the outcome we compared it to our automated evaluation functions (See section 2.5, "Model validation"), to determine if the latter could successfully act as proxies for the former. A generalised linear model using the automated Levenshtein distance and automated F1 score was found to be the best fitting model ($df$ = 24) with an $R^2$ of 0.4276512 and $AUC$ of 0.5957461. This relationship is even stronger when looking just at the amount of training data an $R^2$ of 0.9781961 and $AUC$ of 0.5669595.

As a follow-up, in order to examine the performance of the newly released GPT 4o when compared to that of GPT 3.5, we performed a second fine-tuning process, this time using the model *"gpt-4o-mini"*, analysing only the mean minimum Levenshtein distance of each data point, which we now know to be a good proxy for human interpretability. The fine-tuning also followed a *k*-fold structure, but now performed over a series of 10 (A-J) folds in order to decrease variance. As for the amount of training data, it followed the same structure of sets of 10, 20, 40 and 80 papers.

**FIG S.3.2**. Performance of *"gpt-4o-mini"* at extracting species location data (coordinates and name) calculated as the F1 score (b) and the mean minimum Levenshtein distance (a) between each point from our ground truth dataset and every other point suggested by GPT. The Levenshtein distance is a common metric of dissimilarity between text strings. The fine-tuned models used are trained on different folds (A-J) and differing amounts of data (x-axis). The black line is the mean and standard deviation of all folds.

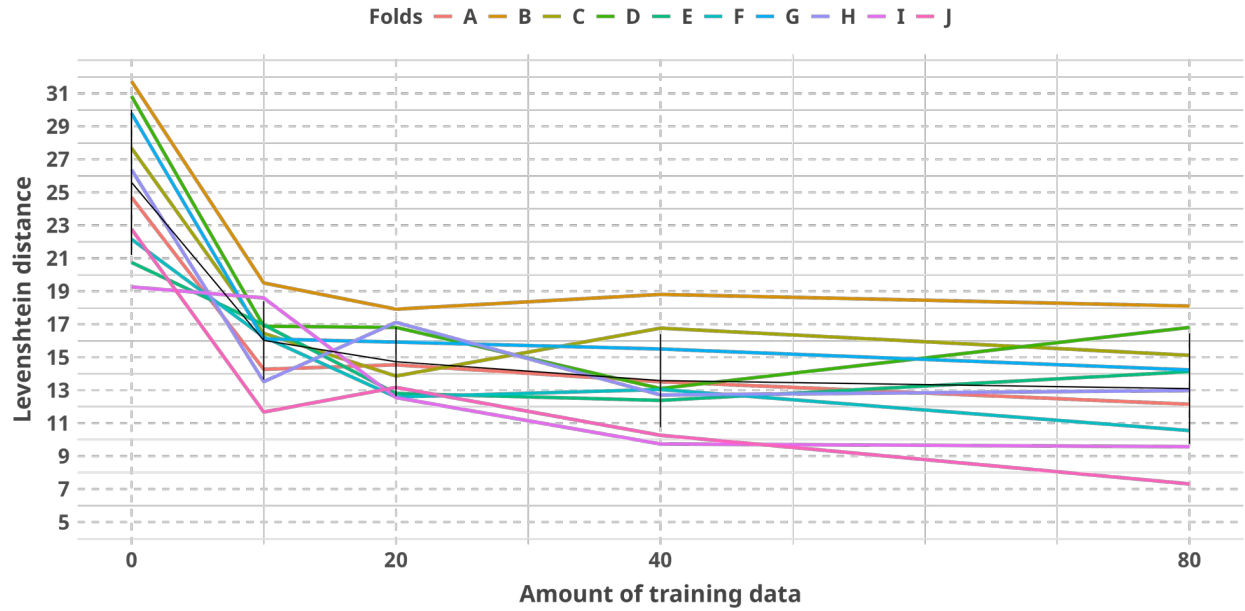

# SUPPORTING INFORMATION - 4

**TABLE S.4.1.** Number of records of each species present in GBIF for a random sample of 100 spider species listed in the World Spider Catalog (WSC). Species not found are omitted.

| Species | GBIF |
| --- | --- |
| *Oxyopes heterophthalmus* | 500 |
| *Dolomedes scriptus* | 500 |
| *Lasaeola prona* | 266 |
| *Socca senicaudata* | 82 |
| *Metellina mimetoides* | 69 |
| *Alopecosa pentheri* | 57 |
| *Pterartoria lativittata* | 46 |
| *Steatoda diamantina* | 40 |
| *Phoroncidia lygeana* | 37 |
| *Staveleya paulae* | 27 |
| Anyphaena rebecae | 23 |
| Tricholathys spiralis | 23 |
| Draconarius pseudospiralis | 22 |
| Anelosimus pratchetti | 14 |
| Euryopis superba | 14 |
| Clubiona littoralis | 14 |
| Pardosa baehrorum | 13 |
| Cyclosa sanctibenedicti | 13 |
| Neon pictus | 12 |
| Carepalxis beelzebub | 11 |

| Species | Count |
|---|---|
| Nipisa phasmoides | 11 |
| Tibellus nigeriensis | 11 |
| Deinopis guianensis | 7 |
| Trochosa arctosina | 6 |
| Noegus actinosus | 6 |
| Metepeira nigriventris | 6 |
| Matinta vicana | 5 |
| Zimiromus jamaicensis | 5 |
| Heliophanus fuerteventurae | 5 |
| Desis japonica | 4 |
| Ameridion signaculum | 4 |
| Alpaida weyrauchi | 4 |
| Heteropoda duan | 4 |
| Dysdera unguimmanis | 4 |
| Anoteropsis canescens | 3 |
| Ozarchaea platnicki | 3 |
| Zelotes cornipalpus | 3 |
| Desognaphosa boolbun | 3 |
| Molycria monteithi | 3 |
| Thunberga woodae | 3 |
| Agelena littoricola | 2 |
| Paraeboria jeniseica | 2 |
| Aelurillus spinicrus | 2 |
| Strotarchus tamaulipas | 1 |
| Negayan tarapaca | 1 |
| Tabuina baiteta | 1 |
| Tarne dives | 1 |
| Myrmecium urcuchillay | 1 |
| Singa hilira | 1 |
| Nesticella sogi | 1 |
| Stasimopus filmeri | 1 |
| Neriene herbosa | 1 |
| Belisana erawan | 1 |
| Philodromus schicki | 1 |
| Clubiona kayashimai | 1 |
| Bathyphantes montanus | 1 |
| Myrmarachne palladia | 1 |
| Antrodiaetus effeminatus | 1 |
| Deltoclita rubra | 1 |
| Ariadna lalen | 1 |

**Table S.4.2**. Number of records of each species extracted by *ARETE* for a random sample of 100 spider species listed in the World Spider Catalog (WSC). Species not found are omitted.

| Species | ARETE |
|---|---|
| Dolomedes scriptus | 67 |
| Alopecosa pentheri | 63 |
| Ariadna lalen | 60 |
| Oxyopes heterophthalmus | 27 |
| Oreoneta kurile | 17 |
| Socca senicaudata | 15 |
| Lasaeola prona | 10 |
| Hexurella zas | 7 |
| Nipisa phasmoides | 7 |
| Alloclubionoides rostratus | 6 |
| Cantikus exceptus | 6 |
| Escaphiella catemaco | 6 |
| Myrmecium urcuchillay | 6 |
| Neriene herbosa | 6 |
| Helsdingenia ceylonica | 5 |
| Clubiona littoralis | 4 |
| Eleleis okavango | 4 |
| Hersilia orvakalensis | 4 |
| Negayan tarapaca | 4 |
| Stasimopus filmeri | 4 |
| Tarne dives | 4 |
| Trichoncoides striganovae | 4 |
| Anoteropsis canescens | 3 |
| Draconarius pseudospiralis | 3 |
| Metepeira nigriventris | 3 |
| Namundra murphyi | 3 |
| Pardosa baehrorum | 3 |
| Tabuina baiteta | 3 |
| Thunberga woodae | 3 |
| Tricholathys spiralis | 3 |
| Zelotes cornipalpus | 3 |
| Draconarius cheni | 2 |
| Alpaida weyrauchi | 2 |
| Anelosimus pratchetti | 2 |
| Clubiona kayashimai | 2 |
| Dysdera unguimmanis | 2 |
| Heliophanus fuerteventurae | 2 |
| Mangora divisor | 2 |
| Matinta vicana | 2 |
| Metellina mimetoides | 2 |
| Myrmarachne palladia | 2 |
| Pimoa phaplu | 2 |
| Strotarchus tamaulipas | 2 |
| Tetragnatha obscura | 2 |
| Zodarion beroni | 2 |
| Ameridion signaculum | 1 |
| Antrodiaetus effeminatus | 1 |
| Anyphaena rebecae | 1 |
| Apollophanes indistinctus | 1 |
| Belisana erawan | 1 |
| Crocodilosa virulenta | 1 |
| Cyclosa punjabiensis | 1 |
| Cyrtophora gazellae | 1 |
| Desognaphosa boolbun | 1 |
| Euryopis superba | 1 |
| Heteropoda duan | 1 |
| Mesabolivar forceps | 1 |
| Nesticella sogi | 1 |
| Noegus actinosus | 1 |
| Orcevia zabkai | 1 |
| Persilena sengleti | 1 |
| Philodromus ketani | 1 |
| Philodromus schicki | 1 |
| Staveleya paulae | 1 |
| Steatoda diamantina | 1 |
| Tibellus nigeriensis | 1 |
| Zimiromus jamaicensis | 1 |

# SUPPORTING INFORMATION - 5

## S.5.1 Advantages

Due to the large dataset sizes expected from users and the complexity of analyzing the similarity between two non-numeric concepts, *ARETE* comes integrated with tools for simplifying the comparison between model and reference data. Function *performance_report()* can take two matrices of equal structure with species, file data and a user's choice of locations, coordinates or both. One matrix is the data output by a LLM, the other is the reference data.

The foremost advantages of LLM extraction of species data are the speed and cost of data extraction. Although currently our pipeline achieves speeds of 3.7 pages/minute which is close but less than an experienced human annotator at a comfortable pace, 4.3 pages/minute, other aspects must be considered. Assuming 8 hours of work at a constant pace a human will process roughly 2070.5 pages, compared to GPT's 5341.4 at the end of the (24 hour) day. Using GPT is also less costly for the user. For a given benchmark test paper of 16 pages we found that an annotator completed it in 3.9 min, against the 4.3 mins of *ARETE*. Assuming an annotator earns a salary of 2700€ for 23 days of 8 daily hours of work (Finnish average salary for a PhD student), their cost would be 0.2€/min. This makes the total cost of data extraction for this paper by an annotator, 1€ compared to *ARETE*'s 0.003$ (at current prices of ChatGPT), making it a nearly 315 times less costly procedure.

We argue that the current costs of running our pipeline are not significant in the context of scientific research. Additionally, it is not strictly necessary to use a paid service in already existing models. ChatGPT currently allows new users to start a free-trial account with a limited amount of resources and limited number of requests per minute using their API. This is useful for prospective researchers as a way to determine if GenAI might fit in their research and as such, *ARETE* offers the option to run on a free account, respecting its current limitations in the total amount of tokens per request and the frequencies of requests.

The other major aspect of *ARETE* that distinguishes it from similar human efforts is its handling of model errors. When considering user supplied reference data as ground truth it is possible to classify the outcome of model predictions and their error types. As such, True Positives are points shared among model and annotator sets, False Positives (*i.e.* false information, Type I errors, or "hallucinations" when produced by LLMs) are points represented in the model set but not in the annotator set and False Negatives are points represented in the annotator set but not in the model set (*i.e.* information not detected, Type II errors). This is crucial as in conservation biology error types differ clearly in their weight. The most damaging type of error overall in conservation research and subsequent decision making are False Positives. These can result, for example, in less conservative estimates for species distributions which potentially lead to deescalating conservation efforts in critical species. The result of a species going extinct due to an underestimation of its extinction risk is much harder to recover from compared to the overzealous protections that could result from False Negatives. Due to the unbalanced impact of error types in conservation work, we have taken care to apply methods that highlight and reduce these errors (see sections 2.4, "Outlier processing" and 2.5, "Model validation"), the success of which can be observed ARETE's in high precision with very few true "hallucinations" (see section 4, "Model performance" for more details on these). Lastly, some types of potential errors are completely absent, such as hallucinating entirely new species names (or data) completely unrelated to the text. While it's completely possible that these could still occur, we're certain that, given our test size, these should be a negligible amount of mistakes in the extraction of most data.

We believe both of these aspects make a compelling argument for many projects to consider the inclusion of *ARETE*. As of now the usage of LLM is, at least, competitive with human annotators under ideal circumstances which will frequently not be present. Additional considerations over the time and cost of training human annotators will further push ARETE and future GenAI alternatives as advantageous over human annotators.

## S.5.2 Limitations & challenges

### S.5.2.1 Resource limited validation

Although the performance achieved in *ARETE* is remarkable we would like to explore further aspects of LLM validation. Doing so can extend its possible application to further kinds of work. As such we are considering a variety of other validation tests that we will continuously make available. This includes, for example, performing both inter-rater reliability and intra-rater reliability analyses (Gisev *et al.*, 2013), to determine the consistency of model output versus those of human annotators. However, we have opted not to include it in this paper as to perform it according to the standards in fields such as psychology, a number of human annotators are needed that surpass our current project funding. Furthermore, there are also questions of performance while using thematically different data, *i.e.*, papers outside of the scope of *RECODE*. Furthermore, some situations will organically appear in taxonomy papers that we did not encounter in our data but know that exist. For example, we did not see any cases of author data dispute in our training data. We do not know how existing LLM will behave in this scenario. In both cases, although we have not encountered anything to suggest that it should have vastly different performance, they remain unvalidated and we encourage users to make use of our automated validation tools and documentation on fine tuning in order to adapt *ARETE* to their needs.

### S.5.2.2 Black box & proprietary nature

Unfortunately, the common proprietary nature of most LLM generates issues of cost and transparency that are difficult to overcome. While we are certain that free, better open-access options will be available in the future, for now the best LLM performance is commercial. In terms of transparency, the black box nature of LLMs is a known issue that makes it impossible to ever ascertain the reasoning behind the data extracted. Unknown biases in the training may hinder performance, and may be difficult to identify especially if the effect is light. Moreover, all fine-tuned models reported in this paper are inaccessible to the public. This is by design: only the creator of the model and the limited members of their team (in some pay plans) may use these models. We believe, however, they are easy to recreate as the dataset we used is available.

### S.5.2.3 Performance & other usage limitations

The current architecture of LLM has at its core neural networks which draw up results through statistical likelihood instead of conceptual understanding, with little evidence to suggest that conceptualization as we understand it could develop spontaneously in hyper-complex systems, i.e., those with upwards of 1 million parameters. As such, a LLM will tend to be more inflexible and more frequently unable to complete concept-dependent tasks which might be necessary to document interpretation, such as those requiring extrapolation (e.g. asking for generation of a clock pointing to a time other than 10:10, an issue still in GPT-4o).

A common problem brought up in the extraction of species data is also the correspondence between the species names extracted and those taxonomically valid. Although right now this is not implemented we are planning on introducing an optional check between model suggestions and online taxonomy databases. This feature will probably be introduced for a reduced number of taxa such as spiders and then expanded later on, as not all taxa have widely accepted online reference databases such as the World Spider Catalog.

Lastly, ARETE so far only makes use and tests the usage of LLM in English written text. This is due to known biases in the data used to pretrain these models. Lack of experience with non-English nouns, may lead to added difficulty due to non identification or misattribution. This could result in a performance loss in papers containing key non-English terms that would go undetected until specifically tested for.

## S.5.3 References

- Gisev, N., Bell, J.S. & Chen, T.F. (2013) 'Interrater Agreement and Interrater Reliability: Key concepts, approaches, and applications', *Research in Social and Administrative Pharmacy*, 9(3), pp. 330–338. doi:10.1016/j.sapharm.2012.04.004.